\documentclass{article}

\usepackage[preprint, nonatbib]{neurips_2020}

\usepackage[utf8]{inputenc} 
\usepackage[T1]{fontenc}    
\usepackage{hyperref}       
\usepackage{url}            
\usepackage{booktabs}       
\usepackage{amsfonts}       
\usepackage{nicefrac}       
\usepackage{microtype}      
\usepackage[minnames=3,maxnames=4,style=ieee,backend=bibtex]{biblatex}
\AtEveryBibitem{
    \clearfield{url}
    \clearfield{urlyear}
    \clearfield{urlmonth}
}

\usepackage{graphicx}
\usepackage{subcaption}
\usepackage{cleveref}
\usepackage[flushleft]{threeparttable}
\usepackage{caption} 
\usepackage[titletoc,page]{appendix}
\usepackage{rotating}
\usepackage{hvfloat}
\usepackage[section]{placeins}
\usepackage{minitoc}
 
\DeclareUnicodeCharacter{FB01}{fi}

\title{Machine-Learning Driven Drug Repurposing for COVID-19}

\author{%
  Semih Cantürk \thanks{Equal contributions} \\
  Zetane Systems \\
  1870 Boulevard des Sources, Suite 307 \\
  Pointe-Claire, Canada H9R 5N4 \\
  \texttt{semih@zetane.com} \\
  \And
  Aman Singh \footnotemark[1]\\
  Zetane Systems \\
  1870 Boulevard des Sources, Suite 307 \\
  Pointe-Claire, Canada H9R 5N4 \\
  \texttt{aman@zetane.com} \\
  \And
  Patrick St-Amant \footnotemark[1]\\
  Zetane Systems \\
  1870 Boulevard des Sources, Suite 307 \\
  Pointe-Claire, Canada H9R 5N4 \\
  \texttt{patrick@zetane.com} \\
  \And
  Jason Behrmann \\
  Zetane Systems \\
  1870 Boulevard des Sources, Suite 307 \\
  Pointe-Claire, Canada H9R 5N4 \\
  \texttt{jason@zetane.com} \\
}

\begin{document}

\maketitle

\begin{abstract}
  The integration of machine learning methods into bioinformatics provides particular benefits in identifying how therapeutics effective in one context might have utility in an unknown clinical context or against a novel pathology. We aim to discover the underlying associations between viral proteins and antiviral therapeutics that are effective against them by employing neural network models. Using the National Center for Biotechnology Information virus protein database and the DrugVirus database, which provides a comprehensive report of broad-spectrum antiviral agents (BSAAs) and viruses they inhibit, we trained ANN models with virus protein sequences as inputs and antiviral agents deemed safe-in-humans as outputs. Model training excluded SARS-CoV-2 proteins and included only Phases II, III, IV and Approved level drugs. Using sequences for SARS-CoV-2 (the coronavirus that causes COVID-19) as inputs to the trained models produces outputs of tentative safe-in-human antiviral candidates for treating COVID-19. Our results suggest multiple drug candidates, some of which complement recent findings from noteworthy clinical studies. Our in-silico approach to drug repurposing has promise in identifying new drug candidates and treatments for other viruses.
\end{abstract}

\section{Introduction}

Artificial intelligence (AI) technology is a recent addition to bioinformatics that shows much promise in streamlining the discovery of pharmacologically active compounds \cite{stephenson_survey_2019}. Machine learning (ML) provides particular benefits in identifying how drugs effective in one context might have utility in an unknown clinical context or against a novel pathology \cite{napolitano_drug_2013}. The application of ML in biomedical research provides new means to conduct exploratory studies and high-throughput analyses using information already available. In addition to deriving more value from past research, researchers can develop ML tools in relatively short periods of time.

Past research now provides a sizable bank of information concerning drug-biomolecule interactions. Using drug repurposing as an example, we can now train predictive algorithms to identify patterns in how antiviral compounds bind to proteins from diverse virus species. We aim to train an ML model so that when presented with the proteome of a novel virus, it will suggest antivirals based on the protein segments present in the proteome. The final output from the model is a best-fit prediction as to which known antivirals are likely to associate with those familiar protein segments.

These benefits are of particular interest for the current COVID-19 health crisis. The novelty of SARS-CoV-2 requires that we execute health interventions based on past observations. Grappling with an unforeseen pandemic with no known treatments or vaccines, the potential for rapid innovation from ML is of utmost significance. The ability to conduct complex analyses with ML enables us to research insights quickly that can help steer us in the right direction for future studies likely to produce fruitful results. We present here multiple models that produced a number of antiviral candidates for treating COVID-19. Out of our 12 top predicted drugs, 6 of them have shown positive results in recent findings based on cell culture results and clinical trials. These promising antivirals are lopinavir, ritonavir, ribavirin \cite{hung_triple_2020}, cyclosporine, \cite{de_wilde_cyclosporin_2011-1}, rapamycin \cite{gordon_sars-cov-2_2020}, and nitazoxanide \cite{wang_remdesivir_2020}. For the 7 other predicted drugs, further research is needed to evaluate their effectiveness against SARS-CoV-2.

\section{Method}

\subsection{Data}

\subsubsection{Sourcing and Preparation}

We used two main data sources for this study. The first database was the DrugVirus database \cite{andersen_discovery_2020}; DrugVirus provides a database of broad-spectrum antiviral agents (BSAAs) and the associated viruses they inhibit. The database covers 83 viruses and 126 compounds, and provides information on the status as antiviral of each compound-virus pair. These statuses fall into eight categories representing the progressive drug trial phases: Cell cultures/co-cultures, Primary cells/ organoids, Animal model, Phases I-IV and Approved. See Appendix A for a more intuitive pivot table view of the database.

The second database is the National Center for Biotechnology Information (NCBI) Virus Portal \cite{brister_ncbi_2015}; as of April 2020, this database provides approximately 2.8 million amino-acid and 2 million nucleotide sequences from viruses with humans as hosts. Each row of this database contains an amino acid sequence specimen from a study, as well as metadata that includes the associated virus species. In our work, we considered sequences only from the 83 virus species in the DrugVirus database or their subspecies in order to be able to merge the two data sources successfully. We also constrained ourselves to amino-acid sequences only in the current iteration. The main reasons for this are two-fold:

\begin{enumerate}
  \item Amino-acid sequences are essentially derived from the DNA sequences, which may encode overlapping information on different levels. In somewhat simplified terms, amino-acid sequences are the outputs of a layer of preprocessing on genetic material (in the form of DNA/RNA).
  \item Nucleotide triplets (codons) map to amino-acids, making amino-acid sequences much shorter and easier to extract features both in preprocessing and in the machine learning methods themselves. Shorter sequences also mean the ML pipeline will be more resource-efficient, i.e. easier to train.
\end{enumerate}

The amino-acids were downloaded as three datasets: HIV types 1 \& 2 (1,192,754 sequences), Influenza types A, B \& C (644,483 sequences), and the “Main” dataset for all other types including SARS-CoV-2 (785,624 sequences). Each dataset came with two components. The “sequence” component is composed on Accession IDs and the amino-acid sequence itself, while the “metadata” component includes all other data (e.g. virus species, date specimen was taken, an identifier of the related study) as well as the Accession ID to enable merging the two components.

The amount of research with a focus on Influenza and HIV naturally lead to these viruses comprising most of the samples. In our experiments, we have excluded these viruses, and have worked only with dataset \#3, though the other datasets can be integrated into the main one during the class balancing process, an idea we will discuss in \Cref{futurework}, Future Work.

\subsubsection{Preprocessing}

The first step of the preparation phase was to merge the “sequence” and “metadata” components into a single NCBI dataset based on sequence IDs. Afterwards, we mapped the "Species" column in this main dataset to the Virus Name column in the DrugVirus database. This step was required as these two columns that denote the virus species in the respective datasets did not match due to subspecies present in the sequence dataset and alternative naming of some viruses.

Afterwards, we processed the DrugVirus dataset to a format suitable for merging with the NCBI data frame. Every row of the DrugVirus dataset consists of a single drug-virus pairing and their respective interaction/drug trial phase, meaning any given drug and virus appeared in multiple rows of the dataset. We derived a new DrugVirus dataset that functioned as a dictionary where each unique virus was a key, and the interactions with antivirals encoded as a multi-label binary vector ($1$ if viable antiviral according to the original dataset, $0$ if not) of length 126 (the number of antivirals) which corresponded to the value. We came up with three “versions” depending on how we decided an antiviral was a viable candidate to inhibit a virus. The criteria depended on drug trial phases:

\begin{enumerate}
    \item In the first version, any interaction between a drug-virus pair is designated by a $1$. This means drugs that did not go past cell cultures/co-cultures or primary cells/organoids testing are still considered viable candidates.
    \item This second version expands upon the first stemming from our discovery that an attained trial phase in the database does not necessarily mean previous phases were also listed in the database. For example, we found that for a given virus, a given drug had undergone Phase III testing, designated by a $1$, but Phase I \& II were listed as $0$s. This undermined our assumption that drug trials are hierarchical; though, in reality this is usually the case. This can be caused by missing data reporting or possibly skipped phases. We proceeded with the hierarchy assumption, and extended the database in (1) to account for the previous phases. This meant that in this second version, an Approved drug will have all phases designated with $1$s, for example. Keeping track of the 8 phases meant that the size of the database also grew by 8.
    \item In the third version, we considered a drug-virus pair as viable only if it has attained Phase II or further drug trials, signifying some success with human trials have been observed. In the results presented in \Cref{results}, our training database was based on this third version of the DrugVirus database.
\end{enumerate}

The full dataset was then generated by merging this “new” version of the DrugVirus dataset with the NCBI dataset. We then generated two versions of this full dataset: one that consists of all SARS-CoV-2 sequences and one that consists of all other viruses available. This enabled us to compare how successful our models are in a case when they have not been trained on the virus species at all and have to detect peptide substructures in the sequences to suggest antivirals. A sample of this final database (with some columns excluded for brevity) is available in Appendix B.

Upon inspection of the data, we found that there were replete of duplicate or extremely similar virus sequences. To reduce this exploitability and pose a more challenging problem, we removed the duplicate sequences that belonged to the same species and had the exact same length. This reduced the size of the dataset by approximately 98\%. The counts for each virus species before and after dropping duplicate viruses is available in Appendix C1 and C2.

\subsubsection{Balancing}

Our main database also contained a class imbalance in the number of times certain virus species appeared in the database. We oversampled rare viruses (e.g., West Nile virus: 175 sequences) and excluded the very rare species which compose less than 0.5\% of the available unique samples in the dataset (e.g., Andes virus: 4 sequences), and undersampled the common viruses (e.g., Hepatitis C: 16,040 sequences). This produced a more modest database of 30,479 amino acid sequences, with each virus having samples in the 400–900 range (see Appendix C3). We kept the size of the dataset small both to enable easier model training and validation in early iterations and to handle data imbalance more smoothly.

The class imbalance problem also presented itself in the antiviral compounds. Even with balanced virus classes, the number of times each drug occurred within the dataset varied, simply because some drugs apply to more viruses than others. To alleviate this, we computed class weights for each drug, which we then provided to the models in training. This enabled a fairer assessment and a more varied distribution of antivirals in predicted outputs.

\subsubsection{Train/Test Splitting}\label{splitting}

The final step of data processing involved generating the training and validation sets. We split the data in two different ways, resulting in two different experiments (see \Cref{expsetup}, Experiment Setup for the full experiment pipeline). \textit{Experiment I} is based on a standard, randomized an 80\% training/20\% validation split on the main dataset.

For \textit{Experiment II}, we split the data on virus species, meaning the models were forced to predict drugs for a species that it was not trained on, and have to detect peptide substructures in the amino-acid sequences to suggest drugs. In this setup we also guaranteed that the SARS-CoV-2 sequences were always in the test set, in addition to three other viruses randomly picked from the dataset. We used a variant of this setup that trains on all virus sequences except SARS-CoV-2 and is validated on SARS-CoV-2 only to generate the results presented in \Cref{results}.

\subsection{Models}
\label{models}
A growing number of studies demonstrate the success of using artificial neural networks (ANN) in evaluating biological sequences in drug repositioning and repurposing \cite{donner_drug_2018}\cite{zeng_deepdr_2019}. Previous work on training neural networks on nucleotide or amino-acid sequences have been successful with recurrent models such as gated recurrent units (GRU), long short-term memory networks (LSTM) and bidirectional LSTMs (biLSTM), as well as 1D convolutions and 2D convolutional neural networks (CNN) \cite{lee_protein_nodate}\cite{hou_deepsf_2018}. We have therefore focused on these network architectures, and conducted our experiments with an LSTM with 1D convolutions and bidirectional layers as well as a CNN. The network architectures are explained briefly below.

\paragraph{LSTM and 1D Convolutions}
For the LSTM, A character-level tokenizer was used to encode the FASTA sequences into vectors consumable by the network. The sequences were then padded with zeros or cut off to a fixed length 500 to maintain a fixed input size. The network architecture consisted of an embedding layer, followed by 1D convolution and bidirectional LSTM layers (each followed by maxpooling), and two fully connected layers. A more detailed architecture diagram is available in Appendix D.

\paragraph{Convolutional Neural Network (CNN)}
For the CNN, the input features were one-hot encoded based on the FASTA alphabet/charset, which assisted in interpretability when examining the 2D input arrays as images. The inputs are also fixed at a length of 500, resulting in 500 x 28 images, where 28 is the number of elements in the FASTA charset. The network architecture consists of four 2D convolutions with filter sizes of 1x28, 2x28, 3x28 and 5x28 respectively, which are maxpooled, concatenated and passed through a fully connected layer. A more detailed architecture diagram is available in Appendix E.

\subsection{Experiment Setup}
\label{expsetup}

The experiments were run on a computer with an 2.7 GHz Intel Broadwell CPU (61 GB RAM) and NVIDIA K80 GPU (12 GB). Both models completed a 20-epoch experiment in 60-90 minutes. One to three training and evaluation runs were made for each setup during model and hyperparameter selections, and ten training and evaluation runs were done to produce the average metrics in \Cref{results}.

The experiments start by determining the model to use and apply the appropriate preprocessing steps mentioned in \Cref{models}. We then proceed with determining the dataset to train and validate on. This part of the experiment setup is more extensively covered in \Cref{splitting}, Train/Test Splitting. We used binary cross entropy (BCE) loss, Adam optimizer, precision, recall and F1-score as metrics since accuracy tends to be an unreliable metric given the class imbalance and the sparse nature of our outputs. After training and validation, predictions were done on the validation set and the results were post-processed for interpretability. 

In post-processing, we applied a threshold to the sigmoid function outputs of the neural network, where we assigned each drug a probability of being a potential antiviral for a given amino acid sequence. After experimenting with different values, we settled on a threshold value of 0.2. Post-processing outputs a list of drugs that were selected along with the respective probabilities for the drugs being “effective” against the virus with the given amino acid sequence. For other hyperparameters involved as well as information on hyperparameter tuning, see Appendix F.

\section{Results}
\label{results}

Here we present the results for the two experiments described in \Cref{splitting}, Train/Test Splitting. The figures and tables presented in this section are based on the LSTM and CNN architectures described in \Cref{models}, which were trained on 128 batch size and 0.01 and 0.001 learning rates respectively for 20 epochs with an Adam optimizer.

\subsection{Experiment I: Train/Test Split}
\begin{figure}
\begin{subfigure}{.5\textwidth}
  \centering
  \includegraphics[width=1\linewidth]{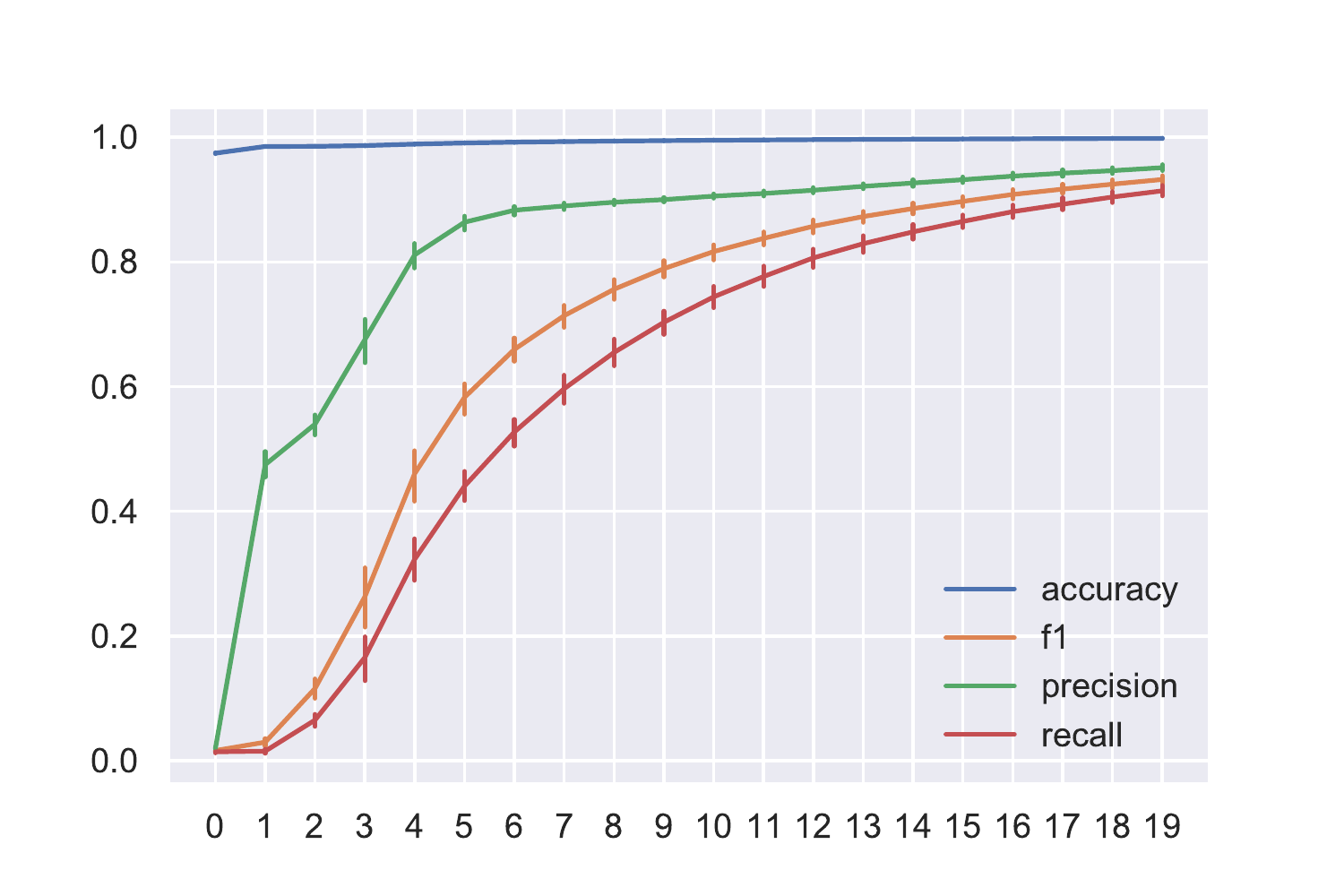}
  \caption{Training set}
  \label{fig:sfig1}
\end{subfigure}%
\begin{subfigure}{.5\textwidth}
  \centering
  \includegraphics[width=1\linewidth]{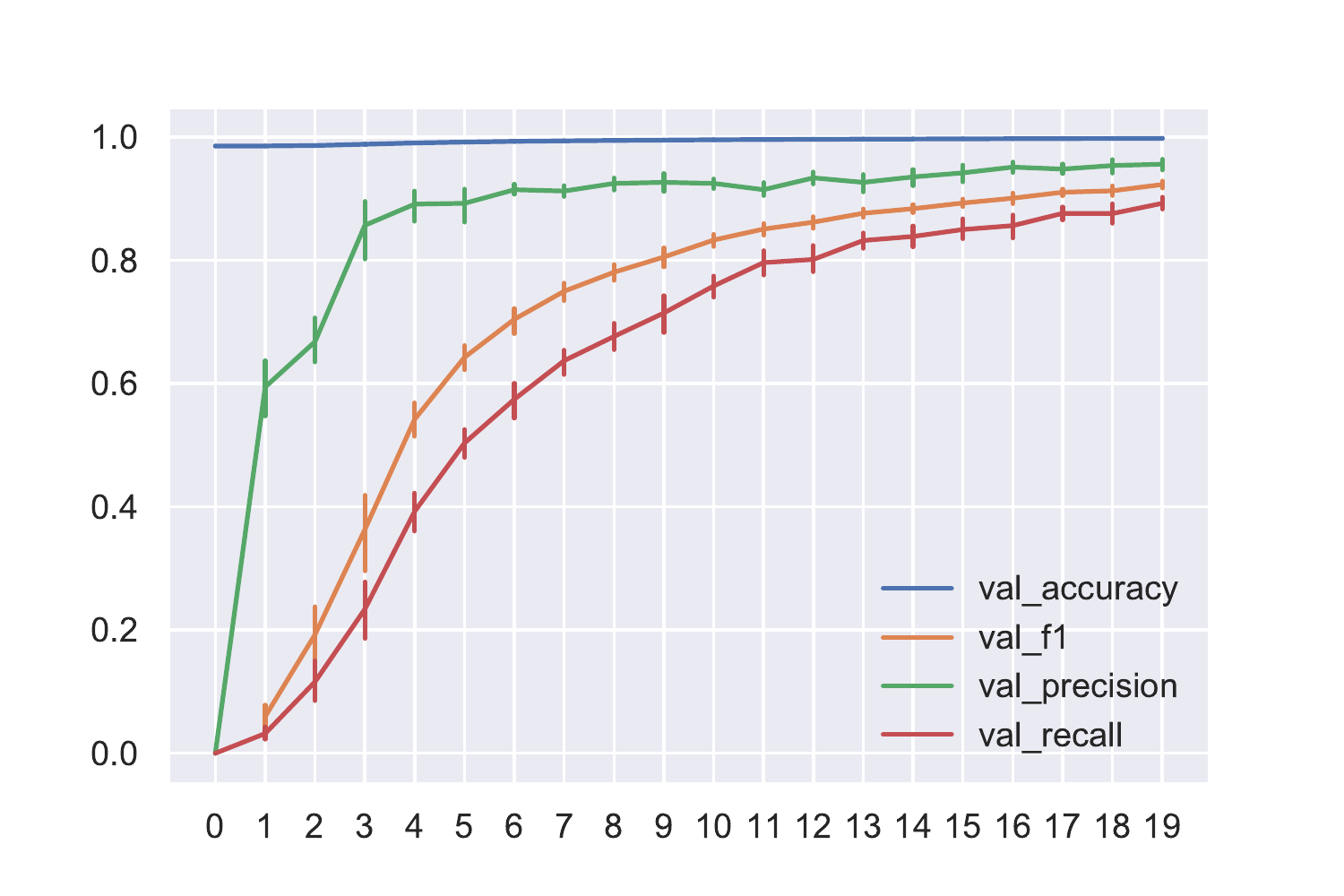}
  \caption{Validation set}
  \label{fig:sfig2}
\end{subfigure}
\caption{Metrics for \textit{Experiment I} over 20 epochs for the LSTM with 95\% confidence intervals, batch size = 128, LR = 0.001}
\label{fig:fig}
\end{figure}

\begin{figure}
\begin{subfigure}{.5\textwidth}
  \centering
  \includegraphics[width=1\linewidth]{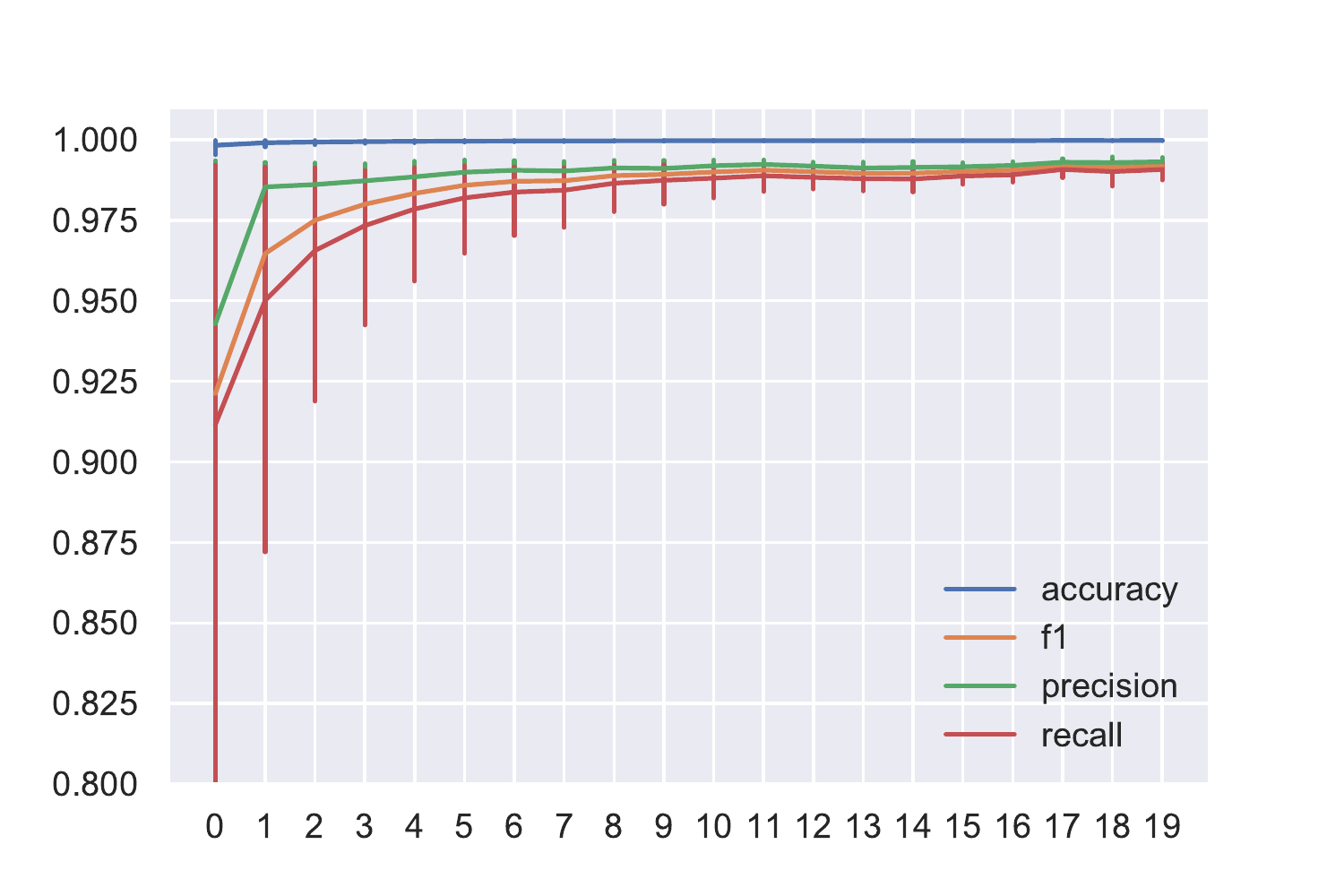}
  \caption{Training set}
  \label{fig2:sfig1}
\end{subfigure}%
\begin{subfigure}{.5\textwidth}
  \centering
  \includegraphics[width=1\linewidth]{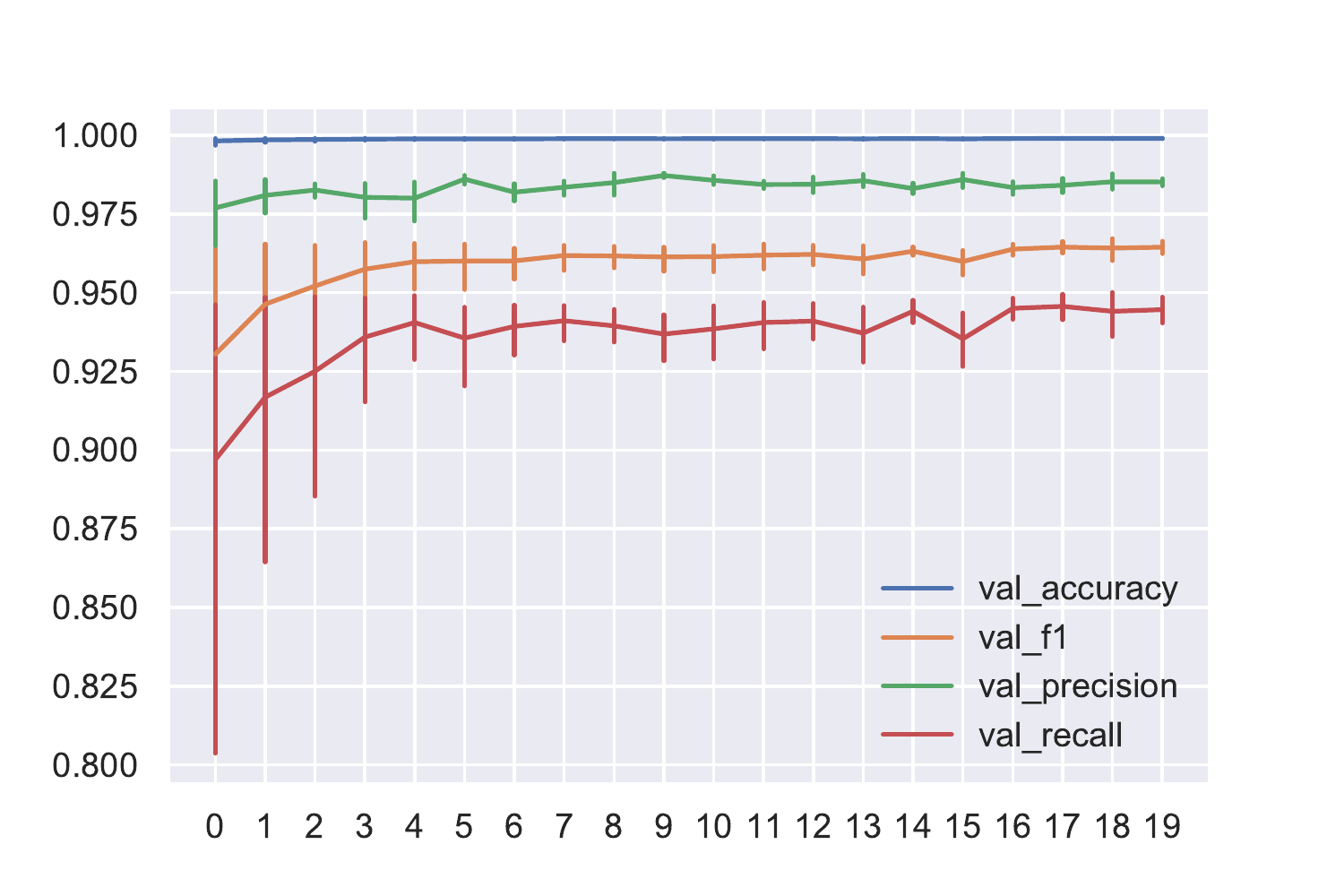}
  \caption{Validation set}
  \label{fig2:sfig2}
\end{subfigure}
\caption{Metrics for \textit{Experiment I} over 20 epochs for the CNN with 95\% confidence intervals, batch size = 128, LR = 0.01}
\label{fig2:fig}
\end{figure}

In the regular setup, we performed an 80\%/20\% train-test split on our data of 30,479 sequences. The metrics for the best set of hyperparameters (based on validation set F1-score) for both the CNN and LSTM architectures respectively are presented in \Cref{table1}. Similarly, plots for the same set of models and hyperparameters over 20 epochs are presented in \Cref{fig:fig,fig2:fig}.

\begin{table}[]
\caption{\textit{Experiment I}, metrics with 95\% confidence intervals for optimal hyperparameters tested.}
\label{table1}
\centering
\begin{tabular}{@{}llllll@{}}
\toprule
\multicolumn{6}{c}{Training set}                           \\ \midrule
Model & Accuracy & Precision & Recall & F1-score & Loss    \\ \midrule
LSTM  & 0.998 $\pm$ 2.3e-4    & 0.951 $\pm$ 0.0059    & 0.914 $\pm$ 0.01  & 0.932 $\pm$ 0.0081   & 0.0126  \\
CNN   & 0.999 $\pm$ 1e-4   & 0.992 $\pm$ 0.0023    & 0.989 $\pm$ 0.0047 & 0.990 $\pm$ 0.0035   & 0.00265 \\ \midrule
\multicolumn{6}{c}{Validation set}                         \\ \midrule
Model & Accuracy & Precision & Recall & F1-score & Loss    \\ \midrule
LSTM  & 0.998 $\pm$ 2.1e-4   & 0.956 $\pm$ 0.011    & 0.892 $\pm$ 0.012  & 0.923 $\pm$ 0.0074   & 0.0134  \\
CNN   & 0.999 $\pm$ 6e-5   & 0.986 $\pm$ 0.0019      & 0.948 $\pm$ 0.0039  & 0.967 $\pm$ 0.0022    & 0.00767  \\ \bottomrule
\end{tabular}
\end{table}

Our models handled the task successfully, achieving 0.958 F1-score in a multi-label multi-class problem setting. This means that the models were able to match the virus species with the sequence substructures and appropriately assign the inhibiting antivirals with accuracy. These satisfactory results led to us implementing \textit{Experiment II}.

\subsection{Experiment II: Predictions on Unseen Virus Species}
\label{experimentii}
In \textit{Experiment II}, the models predicted antiviral drugs for virus species they haven’t been trained on. This meant the models were not able to recommend drugs by “recognizing” the virus from the sequence and therefore had to rely only on peptide substructures in the sequences to assign drugs. In the results presented below, the test set consists of SARS-CoV-2, Herpes simplex virus 1, Human Astrovirus and Ebola virus, whose sequences were removed from the training set.

\begin{table}[]
\caption{\textit{Experiment II}, metrics with 95\% confidence intervals for optimal hyperparameters tested.}
\label{table2}
\centering
\begin{tabular}{@{}llllll@{}}
\toprule
\multicolumn{6}{c}{Training set}                           \\ \midrule
Model & Accuracy & Precision & Recall & F1-score & Loss    \\ \midrule
LSTM  & 0.998 $\pm$  5e-4  & 0.942 $\pm$  0.011  & 0.877 $\pm$ 0.029 & 0.908 $\pm$ 0.021  & 0.00369 \\
CNN   & 0.999 $\pm$ 1.6e-4   & 0.990 $\pm$ 0.0038  & 0.983 $\pm$ 0.0085 & 0.986 $\pm$ 0.0063   & 0.00325 \\ \midrule
\multicolumn{6}{c}{Validation set}                         \\ \midrule
Model & Accuracy & Precision & Recall & F1-score & Loss    \\ \midrule
LSTM  & 0.958 $\pm$ 0.0019   & 0.530 $\pm$ 0.048  & 0.227 $\pm$ 0.014 & 0.318 $\pm$ 0.0099  & 3.764 \\
CNN   & 0.960 $\pm$ 7.9e-4  & 0.588 $\pm$ 0.026  & 0.239 $\pm$ 0.0068 & 0.340 $\pm$ 0.0033 & 7.562\\ \bottomrule
\end{tabular}
\end{table}

\begin{figure}
\begin{subfigure}{.5\textwidth}
  \centering
  \includegraphics[width=1\linewidth]{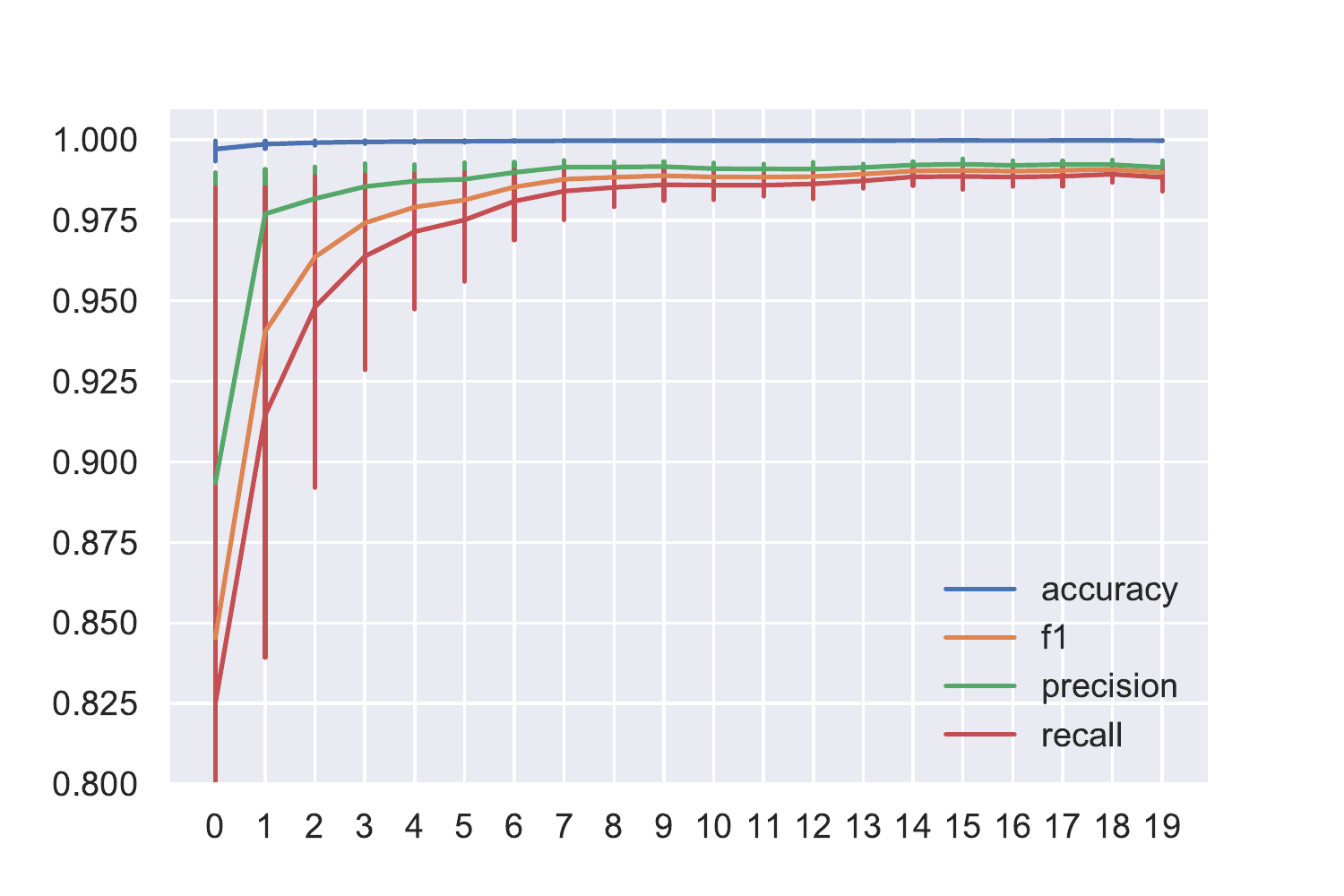}
  \caption{Training set}
  \label{fig3:sfig1}
\end{subfigure}%
\begin{subfigure}{.5\textwidth}
  \centering
  \includegraphics[width=1\linewidth]{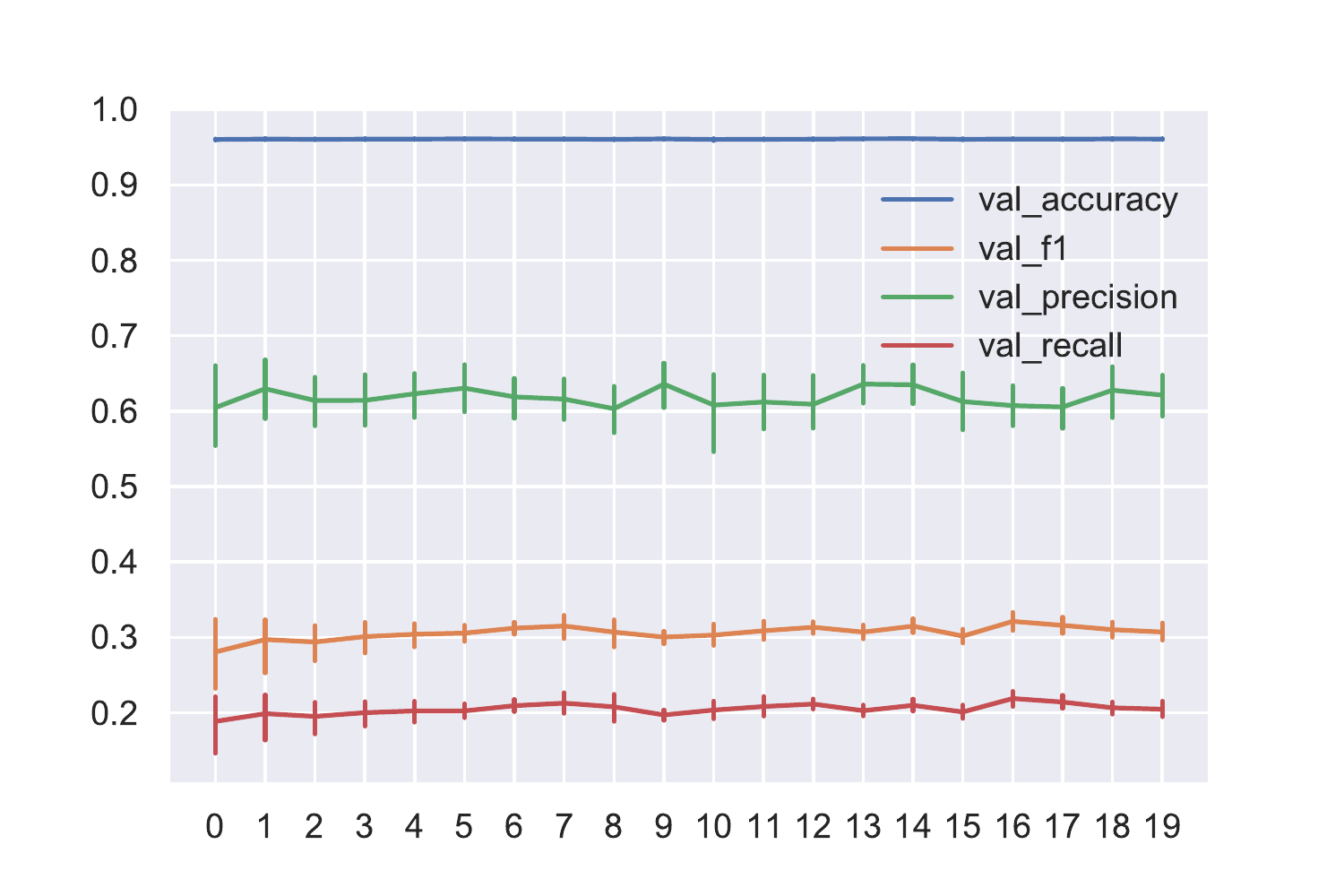}
  \caption{Validation set}
  \label{fig3:sfig2}
\end{subfigure}
\caption{Metrics for \textit{Experiment II} over 20 epochs for the CNN with 95\% confidence intervals, batch size = 128, LR = 0.01}
\label{fig3:fig}
\end{figure}

\begin{table}[]

\caption{\textit{Experiment II} with CNN, Count/Mean Probability table for 588 \textbf{HSV-1} sequences}
\centering
\begin{threeparttable}
\label{table3}

\begin{tabular}{@{}lll@{}}\toprule
Antiviral & Count & Mean Probability \\ 
\midrule
\textbf{Valacyclovir}  & \textbf{578} & \textbf{0.899 $\pm$ 0.0047}\\
\textbf{Cidofovir}   & \textbf{567}    & \textbf{0.737 $\pm$ 0.017}\\ 
\textbf{Foscarnet}   & \textbf{566}    & \textbf{0.904 $\pm$ 0.0066}\\ 
\textbf{Brincidofovir}   & \textbf{559}    & \textbf{0.646 $\pm$ 0.048}\\ 
\textbf{Aciclovir}   &  \textbf{525}   & \textbf{0.759 $\pm$ 0.0081}\\ 
Rapamycin   & 419    & 0.540 $\pm$ 0.017\\ 
Ganciclovir   & 410    & 0.552 $\pm$ 0.064\\ 
Artesunate   & 333    & 0.453 $\pm$ 0.026\\
Cyclosporine   & 328    & 0.449 $\pm$ 0.094\\
Letermovir   & 317    & 0.430 $\pm$ 0.02\\
\textbf{Tilorone}   & \textbf{243}    & \textbf{0.433 $\pm$ 0.071}\\ 
Tenofovir   & 173    & 0.573 $\pm$ 0.017\\ 
\bottomrule
\end{tabular}
\begin{tablenotes}
\item Drugs with more than 100 counts are listed. Antivirals used for Phase II and further trials for HSV-1 are highlighted in \textbf{bold}. Mean probabilities are provided with 95\% confidence intervals.
\end{tablenotes}
\end{threeparttable}
\end{table}

We see here that the CNN (and the LSTM) had issues with convergence, and the accuracies are clearly below their counterparts in the regular setup, though this is certainly expected. We now turn to the actual predictions on the sequences and attempt to interpret them.

Upon examination of drug predictions for Herpes simplex virus 1 (HSV-1), however, we see that our CNN was in fact quite successful. In \Cref{table3} and \Cref{table4}, \textit{Count} represents how many times each drug was flagged as potentially effective for HSV-1 sequences, and \textit{Mean Probability} denotes the average confidence predicted over all instances of the drug. A sample of the outputs where these metrics are derived from is available in Appendix G. Antivirals used for Phase II and further trials for HSV-1 are highlighted in bold, meaning all six drugs in the database that are used for Phase II and further trials are predicted by our model. Three of the top five predictions are approved antivirals for HSV-1 and the only remaining one is predicted 11th among 126 antivirals. This high level of accuracy is remarkable given that our model has not been trained on HSV-1 sequences.

\paragraph{Predictions for SARS-CoV-2}

With some variation between the two, both the LSTM (\Cref{table4a}) and the CNN (\Cref{table4b}) seem to converge on a number of drugs: ritonavir, lopinavir (both Phase III for MERS-CoV), tilorone (Approved for MERS-CoV) and brincidofovir are in the top five candidates in both, while valacyclovir, ganciclovir, rapamycin and cidofovir rank high up in both lists. Most of the remaining drugs are present in both lists as well. The LSTM is more conservative in its predictions than the CNN, and the overall counts for SARS-CoV-2 are significantly lower than for Herpes simplex virus 1 for both, pointing a comparable lack of confidence on the models’ part in predicting SARS-CoV-2 sequences.

A further step we took for the SARS-CoV-2 sequences was visualizing the layer activations in the Zetane Engine to validate that the model was processing the data at a fine-grained level. This was done in similar fashion to a study where integrated gradients were used to generate attributions on a neural network performing molecule classification \cite{mccloskey_using_2019}. The layer activations in both models showed that different antivirals activated different subsequences of a given sequence at the amino acid level, thus validating our approach. The filter activations are available in Appendix H.

\begin{table}
\caption{\textit{Experiment II}, Count/Mean Probability table for 97 \textbf{SARS-CoV-2} sequences, compounded results of 10 trials. Top 20 antivirals shown.}\label{table4}
\begin{subtable}{.5\linewidth}\centering
{\begin{tabular}{@{}lll@{}}
\toprule
Antiviral & Count & Mean Probability \\ 
\midrule
Tilorone   & 321    & 0.627 $\pm$ 0.029\\ 
Brincidofovir   & 197    & 0.457 $\pm$ 0.023\\ 
Ganciclovir   & 113    & 0.469 $\pm$ 0.040\\
Ritonavir   & 106  & 0.589 $\pm$ 0.049\\
Lopinavir   & 106     & 0.588 $\pm$ 0.049\\
Valacyclovir  & 93    & 0.522 $\pm$ 0.046\\
Cidofovir   & 83    & 0.400 $\pm$ 0.039\\ 
Foscarnet   &  58   & 0.473 $\pm$ 0.047\\
Artesunate   & 55    & 0.385 $\pm$ 0.048\\ 
Nitazoxanide   & 48     & 0.359 $\pm$ 0.048\\
Rapamycin   & 32    & 0.399 $\pm$ 0.065\\
Pleconaril   & 30      & 0.626 $\pm$ 0.059\\
Cyclosporine   & 24    & 0.429 $\pm$ 0.084\\ 
Chloroquine   & 21   & 0.499 $\pm$ 0.11\\
Letermovir   & 19      & 0.395 $\pm$ 0.089\\
Aciclovir   & 19      & 0.347 $\pm$ 0.077\\
Ribavirin   & 17      & 0.415 $\pm$ 0.10\\
CYT107  & 5      & 0.369 $\pm$ 0.39\\
Tenofovir   & 5      & 0.340 $\pm$ 0.20\\
Alisporivir   & 3      & 0.550 $\pm$ 0.80\\
\bottomrule
\end{tabular}}
\caption{LSTM}\label{table4a}
\end{subtable}%
\begin{subtable}{.5\linewidth}\centering
{\begin{tabular}{@{}lll@{}}
\toprule
Antiviral & Count & Mean Probability \\ 
\midrule
Tilorone   & 177    & 0.885 $\pm$ 0.031\\ 
Brincidofovir   & 116    & 0.807 $\pm$ 0.044\\ 
Ritonavir   & 114  & 0.808 $\pm$ 0.046\\
Lopinavir   & 93     & 0.881 $\pm$ 0.030\\
Valacyclovir  & 35    & 0.748 $\pm$ 0.085\\
Ganciclovir   & 31    & 0.793 $\pm$ 0.089\\
Cidofovir   & 26    & 0.694 $\pm$ 0.10\\ 
Nitazoxanide   & 24     & 0.767 $\pm$ 0.13\\
Artesunate   & 23    & 0.740 $\pm$ 0.10\\ 
Rapamycin   & 12    & 0.769 $\pm$ 0.14\\ 
Foscarnet   &  18   & 0.809 $\pm$ 0.11\\
Cyclosporine   & 10    & 0.780 $\pm$ 0.22\\ 
Ribavirin   & 9      & 0.731 $\pm$ 0.23\\
Aciclovir   & 8      & 0.671 $\pm$ 0.20\\
Chloroquine   & 6   & 0.938 $\pm$ 0.064\\
Thymalfasin & 5      & 0.826 $\pm$ 0.38\\
Azithromycin & 5      & 0.648 $\pm$ 0.49\\
CYT107  & 3      & 0.540 $\pm$ 0.94\\
Tenofovir   & 2      & 0.990 $\pm$ 0.11\\
Letermovir   & 2      & 0.550 $\pm$ 0.88\\

\bottomrule
\end{tabular}}
\caption{CNN}\label{table4b}
\end{subtable}
\end{table}

\section{Discussion and Future Work}
\label{futurework}

The preliminary results of our experiments show promise and merit further investigation. We note that our ML models predict that some antivirals that show promise as treatments against MERS-CoV may also be effective against SARS-CoV-2. These include the broad-spectrum antiviral tilorone \cite{ekins_tilorone_2020} and the drug lopinavir \cite{yao_systematic_2020}, the latter of which is now in Phase IV clinical trials to determine its efficacy against COVID-19 \cite{basha_corona_nodate}. Such observations suggest with confidence that our models can recognize reliable patterns between particular antivirals and species of viruses containing homologous amino acid sequences in their proteome.

Additional observations that support our findings have come to light from a study in \textit{The Lancet} published shortly before this article \cite{hung_triple_2020}. This open-label, randomized, Phase II trial observed that the combined administration of the drugs interferon beta-1b, lopinavir, ritonavir and ribavirin provides an effective treatment of COVID-19 in patients with mild to moderate symptoms. Both of our models flagged three of the drugs in that trial (note that interferon was not part of our datasets). In terms of number of occurrences aka \textit{Count}, ritonavir, lopinavir and ribavirin were ranked 4th, 5th and 11th by the LSTM, while the CNN model ranked them 3rd, 4th and 10th, respectively.

Other studies have also focused on the treatment of SARS-CoV-2 by drugs predicted in our experiments. Wang et al. discovered that nitazoxanide (LSTM rank 10th, CNN rank 8th) inhibited SARS-CoV-2 at a low-micromolar concentration \cite{wang_remdesivir_2020}. Gordon et al. suggest promising results in viral growth and cytotoxicity assays with rapamycin as an inhibitor (11th \& 10th) \cite{gordon_sars-cov-2_2020}, while de Wilde et al. point out that cyclosporine (13th \& 12th) is known to be effective against diverse coronaviruses \cite{de_wilde_cyclosporin_2011-1}. 

Such observations are encouraging. They demonstrate that predictive models may have value in identifying potential therapeutics that merit priority for advanced clinical trials. They also add to growing observations that support using ML to streamline drug discovery. From that perspective, our models suggest that the broad spectrum antiviral tilorone, for instance, may be a top candidate for COVID-19 clinical trials in the near future. Other candidates highlighted by our results and may merit further studies are brincidofovir, foscarnet, artesunate, cidofovir, valacyclovir and ganciclovir. 

The antivirals identified here have some discrepancies with emerging research findings as well. For instance, our models did not highlight the widely available anti-parasitic ivermectin. One research study observed that ivermectin could inhibit the replication of SARS-CoV-2 \textit{in vitro} \cite{caly_fda-approved_2020}. Another large-scale drug repositioning survey screened a library of nearly 12,000 drugs and identified six candidate antivirals for SARS-CoV-2: PIKfyve kinase inhibitor Apilimod, cysteine protease inhibitors MDL-28170, Z LVG CHN2, VBY-825, and ONO 5334, and the CCR1 antagonist MLN-3897 \cite{riva_large-scale_2020}. It comes as no surprise that our models did not identify these compounds as our data sources did not contain them. Future efforts to strengthen our ML models will thus require us to integrate a growing bank of novel data from emerging research findings into our ML pipeline.

In terms of our machine learning models, better feature extraction can improve predictions drastically. This step involves improvements through better data engineering and working with domain experts who are familiar with applied bioinformatics to better understand the nature of our data and find ways to improve our data processing pipeline. Some proposals for future work that could strengthen the performance of our machine learning process are as follows:

\begin{enumerate}
    \item Deeper interaction with domain experts and further lab testing would lead to a better understanding of the antivirals and the amino-acid sequences they target, leading to building better ML pipelines for drug repurposing.
    \item Better handling of duplicates can improve the quality of data available. The current approach (which is based on species and sequence length) can be improved through using string similarity measures such as Dice coefficient, cosine similarity, Levenshtein distance etc.
    \item Influenza and HIV datasets should be integrated into the data generation and processing pipeline to enhance available data.
    \item Vectorizers can be used to extract features as n-grams (small sequences of chars), which has attained success in similar problems \cite{szalkai_near_2017}. Other unsupervised learning methods such as singular value decomposition also may be applicable to our study \cite{wu_neural_1995}.
\end{enumerate}

\section*{Broader Impact}

We hope that the machine learning approaches and pipelines developed here may provide long-term benefit to public health. The fact that our results show much promise in streamlining drug discovery for SARS-CoV-2 motivates us to adapt our current models so we can conduct identical drug repurposing assessments for other known viruses. Moreover, experimental data suggests that our approaches are generalizable to other viruses (see the HSV-1 example in \Cref{experimentii}, Experiment II) - we are therefore confident that we could adapt our models to conduct equivalent studies during the next outbreak of a novel virus. This also means our methods can be used to repurpose existing drugs in order to find more potent treatments for known viruses.

The direct beneficiaries of our findings are members of the clinical research community. Using relatively few resources, ML-guided drug repurposing technology can help prioritize clinical investigations and streamline drug discovery. In addition to reducing costs and expediting clinical innovation, such efficiency gains may reduce the number of clinical trials -- and thus human subjects used in risky research -- needed to find effective treatments (this pertains to the ethical imperative to avoid harm when possible). Also of importance is that in-silico analyses using machine learning provide yet another means to employ past research findings in new investigations. ML-guided drug repurposing thus provides means to obtain further value from knowledge on-hand; maximizing value in this case is laudable on many fronts, especially in terms of providing maximum benefit from publicly-funded research. 

The negative consequences that could arise should our models fail appear limited but are noteworthy nonetheless. Note that our models aim to only indicate possible therapeutics that merit further clinical investigation in order to prove any antiviral activity against SARS-CoV-2. Should our models fail by recommending spurious treatments, these incorrect predictions may divert limited time and resources towards frivolous investigations. It should also be noted that our methods aim to primarily work as guidance for medical experts, and not as a be-all-end-all solution. And any incorrect inferences made by our models are bound to be detected early by medical experts.

Communicating any machine-learning predictions of tentative antiviral drugs from this study requires much caution. The current pandemic continues to demonstrate how fear, misinformation and a lack of knowledge about a novel communicable disease can encourage counterproductive health-seeking behaviour amongst the public. Soon after the coronavirus became a widely understood threat, the internet was awash in false -- sometimes downright harmful -- information about preventing and treating COVID-19. Included within this misleading health information were premature claims by some prominent government officials that therapeutics like chloroquine and hydroxychloroquine might hold promise as a repurposed drug for COVID-19. Such unfounded advice caused avoidable poisonings from people self-medicating with chloroquine. Subsequent clinical investigations demonstrated no notable benefit and potential adverse reactions to chloroquine when used to treat COVID-19. Such unfortunate events remind us that preliminary findings may be misinterpreted as conclusive treatments or as evidence to support inconclusive health claims.

\begin{ack}
We would like to thank the administrators of the DrugVirus and the NCBI Virus Portal for providing the datasets that are central to this study. We appreciate comments on preliminary drafts of this manuscript from Dr Tariq Daouda from the Massachusetts General Hospital, Broad Institute, Harvard Medical school.

The authors declare they will not obtain any direct financial benefit from investigating and reporting on any given pharmaceutical compound. The following study is funded by the authors’ employer, Zetane Systems, which produces software for AI technologies implemented in industrial and enterprise contexts.
\end{ack}

\printbibliography
\doparttoc
\faketableofcontents 
 

\clearpage

\begin{appendices}
\part{} 
\parttoc

\clearpage

\section{The DrugVirus database displayed in a pivot table}

\begin{figure}[!ht]
\centering
\includegraphics[width=0.98\linewidth]{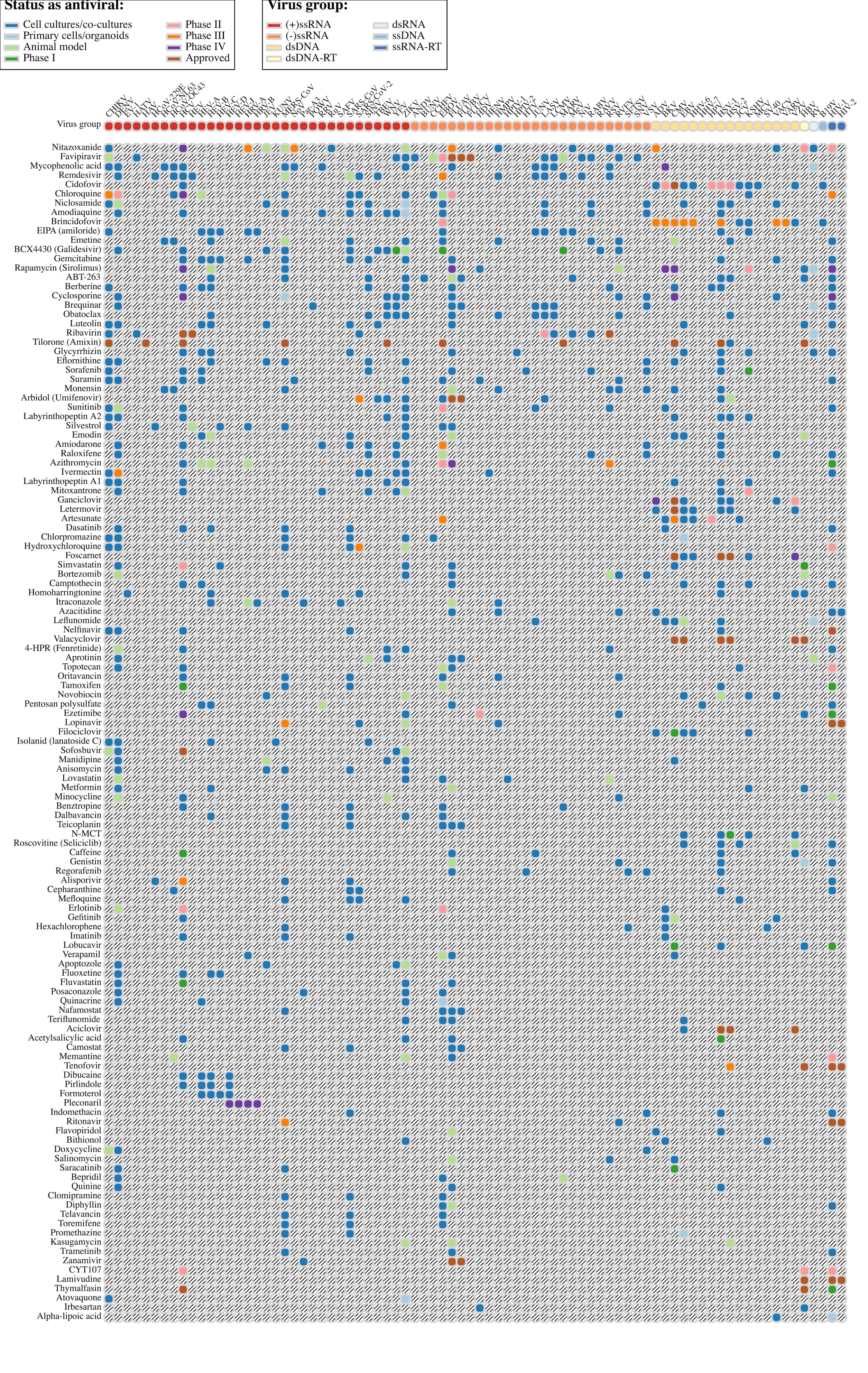}
\end{figure}
  
\section{Merged database sample}

\hvFloat[%
nonFloat=true,
objectAngle=90,%
capPos=r,%
capAngle=90,
capWidth=h,
floatCapSep=10]{table}{
\begin{tabular}{@{}lllcccc@{}}
\toprule
FASTA Sequence & Virus Name                  & GenBank\_Title          & 4-HPR & ABT-263 & Aciclovir & Alisporivir \\ \midrule
YIDPVVVLDF     & Varicella zoster virus      & DNA polymerase catalytic subunit, partial          & 0     & 0       & 1         & 0           \\
YIDPVVVLDF     & Varicella zoster virus      & DNA polymerase catalytic subunit, partial          & 0     & 0       & 1         & 0           \\
TGFYIDPVVV     & Varicella zoster virus      & DNA polymerase catalytic subunit, partial          & 0     & 0       & 1         & 0           \\
ANYNHSDEYL     & Hepatitis A virus           & VP1, partial            & 0     & 0       & 0         & 0           \\
STTQPQLQTT     & Herpes simplex virus 1      & glycoprotein G, partial & 0     & 0       & 1         & 0           \\
NDPITNAVES     & Enterovirus A               & VP1, partial            & 0     & 0       & 0         & 0           \\
MGAQVSTQKT     & Enterovirus B               & polyprotein, partial    & 0     & 0       & 0         & 0           \\
IFEKHSRYKY     & Norovirus                   & RNA replicase, partial  & 0     & 0       & 0         & 0           \\
GCSFSIFLLA     & Hepatitis C virus           & polyprotein, partial    & 0     & 0       & 0         & 1           \\
KKDLKPQTT      & Respiratory syncytial virus & glycoprotein, partial   & 0     & 0       & 0         & 0           \\ \bottomrule
\end{tabular}
}{  A section of the merged database used in model training. Each sequence is associated with a virus species and paired with a length-126 binary vector of antiviral drugs (only 4 elements of the vector are visible), where a $1$ denotes effectiveness linked with a given amino acid sequence. Only the first 10 elements of the amino acid sequence are displayed for brevity.}{mergeddb}
\clearpage

\section{Database profile}
\subsection{Virus counts before dropping duplicate sequences}
\vspace*{\fill}
\begin{table}[!htbp]
\centering
\caption{Virus counts before dropping duplicate sequences}
\begin{subtable}[t]{.5\linewidth}\centering
{
\begin{tabular}[t]{@{}ll@{}}
\toprule
Virus Name                                        & Count \\ \midrule
Hepatitis C virus                                 & 29620 \\
Epstein-Barr virus                                & 16040 \\
Cytomegalovirus                                   & 13089 \\
Norovirus                                         & 7924  \\
Dengue virus                                      & 7720  \\
Hepatitis B virus                                 & 7642  \\
Herpes simplex virus 1                            & 7039  \\
Ebola virus                                       & 6250  \\
Respiratory syncytial virus                       & 6067  \\
Varicella zoster virus                            & 6061  \\
Herpes simplex virus 2                            & 5890  \\
Human papillomavirus                              & 4613  \\
Adenovirus                                        & 4394  \\
Human herpesvirus 6                               & 3622  \\
Human parainfluenza virus 2                       & 2518  \\
Enterovirus A                                     & 2511  \\
Chikungunya virus                                 & 2105  \\
Measles virus                                     & 1672  \\
Lassa virus                                       & 1222  \\
MERS coronavirus      & 1171  \\
Enterovirus C                                     & 1095  \\
Enterovirus D                                     & 1063  \\
SARS coronavirus     & 867   \\
Zika virus                                        & 845   \\
HHV-8           & 838   \\
SFTS virus & 835   \\
Enterovirus B                                     & 831   \\
Human coronavirus OC43                            & 795   \\
Human metapneumovirus                             & 761   \\
Hepatitis E virus                                 & 694   \\
Human parainfluenza virus 1                       & 591   \\
Molluscum contagiosum virus                       & 491   \\
\bottomrule
\end{tabular}
}
\end{subtable}%
\begin{subtable}[t]{.5\linewidth}\centering
{
\begin{tabular}[t]{@{}ll@{}}
\toprule
Virus Name                                        & Count \\ \midrule
Parvovirus B19                                    & 435   \\
Human coronavirus strain NL63                     & 317   \\
Human rhinovirus A                               & 295   \\
Parechovirus A3                                   & 289   \\
Crimean-Congo hemorrhagic fever virus             & 238   \\
Rabies virus                                      & 220   \\
Yellow fever virus                                & 198   \\
Ross River virus                                  & 189   \\
West Nile virus                                   & 175   \\
Marburg virus                                     & 168   \\
Human coronavirus strain 229E                     & 156   \\
Japanese encephalitis virus                       & 153   \\
Hepatitis A virus                                 & 149   \\
Rubella virus                                     & 131   \\
Human rhinovirusÂ B                               & 110   \\
Vaccinia virus                                    & 107   \\
Human herpesvirus 7                               & 107   \\
Rift Valley fever virus                           & 100   \\
Nipah virus                                       & 81    \\
Hantavirus                                        & 76    \\
Tick-borne encephalitis virus                     & 76    \\
John Cunningham virus                             & 50    \\
Saffold virus                                     & 49    \\
Human Astrovirus                                  & 39    \\
Junin virus                                       & 28    \\
LCM virus       & 21    \\
Sindbis virus                                     & 12    \\
Bunyamwera virus                                  & 8     \\
Powassan virus                                    & 5     \\
Andes virus                                       & 4     \\
Sin Nombre virus                                  & 3     \\ \bottomrule
\end{tabular}
}
\end{subtable}
\end{table}
\vspace*{\fill}
\clearpage

\subsection{Virus counts after dropping duplicate sequences}
\vspace*{\fill}
\begin{table}[!htbp]
\centering
\caption{Virus counts after dropping duplicate sequences}
\begin{subtable}[t]{.5\linewidth}\centering
{
\begin{tabular}[t]{@{}ll@{}}
\toprule
Virus Name                                        & Count \\ \midrule
Hepatitis C virus                                 & 870   \\
Dengue virus                                      & 691   \\
Adenovirus                                        & 657   \\
Cytomegalovirus                                   & 654   \\
Hepatitis B virus                                 & 644   \\
Herpes simplex virus 2                            & 635   \\
Herpes simplex virus 1                            & 588   \\
Norovirus                                         & 579   \\
Epstein-Barr virus                                & 545   \\
Respiratory syncytial virus                       & 474   \\
Enterovirus B                                     & 468   \\
Enterovirus A                                     & 417   \\
Human papillomavirus                              & 386   \\
Chikungunya virus                                 & 358   \\
Vaccinia virus                                    & 348   \\
Human herpesvirus 6                               & 322   \\
Enterovirus D                                     & 301   \\
Hepatitis E virus                                 & 293   \\
Human rhinovirus A                                & 289   \\
Human metapneumovirus                             & 284   \\
Molluscum contagiosum virus                       & 284   \\
Enterovirus C                                     & 271   \\
HHV-8           & 264   \\
Parechovirus A3                                   & 247   \\
Zika virus                                        & 223   \\
Varicella zoster virus                            & 213   \\
MERS coronavirus      & 211   \\
Hepatitis A virus                                 & 207   \\
Parvovirus B19                                    & 200   \\
Lassa virus                                       & 188   \\
Ebola virus                                       & 164   \\
Hantavirus                                        & 162   \\
Human parainfluenza virus 2                       & 149   \\
\bottomrule
\end{tabular}
}
\end{subtable}%
\begin{subtable}[t]{.5\linewidth}\centering
{
\begin{tabular}[t]{@{}ll@{}}
\toprule
Virus Name                                        & Count \\ \midrule
Human rhinovirus B                                & 134   \\
Human Astrovirus                                  & 132   \\
Measles virus                                     & 120   \\
Crimean-Congo hemorrhagic fever virus             & 118   \\
Saffold virus                                     & 112   \\
Human coronavirus OC43                            & 104   \\
Human coronavirus strain NL63                     & 87    \\
Human herpesvirus 7                               & 87    \\
John Cunningham virus                             & 83    \\
Rabies virus                                      & 81    \\
SFTS virus & 79    \\
West Nile virus                                   & 75    \\
Rubella virus                                     & 73    \\
Human parainfluenza virus 1                       & 73    \\
Hepatitis D virus                                 & 65    \\
Human coronavirus strain 229E                     & 57    \\
Rift Valley fever virus                           & 47    \\
Yellow fever virus                                & 40    \\
Tick-borne encephalitis virus                     & 36    \\
Japanese encephalitis virus                       & 32    \\
Nipah virus                                       & 29    \\
SARS coronavirus     & 22    \\
Andes virus                                       & 20    \\
Junin virus                                       & 15    \\
LCM virus & 15    \\
Ross River virus                                  & 12    \\
Marburg virus                                     & 12    \\
Sin Nombre virus                                  & 11    \\
Sindbis virus                                     & 4     \\
Bunyamwera virus                                  & 4     \\
Powassan virus                                    & 4     \\
BK virus                                          & 1     \\ \bottomrule
\end{tabular}
}
\end{subtable}
\end{table}
\vspace*{\fill}
\clearpage

\subsection{Virus counts after dropping duplicate sequences and balancing}
\vspace*{\fill}
\begin{table}[h]
\centering
\caption{Virus counts after dropping duplicate sequences and balancing}
\begin{tabular}{@{}ll@{}}
\toprule
Virus Name                                        & Count \\ \midrule
Enterovirus B                                     & 936 \\
Hepatitis C virus                                 & 870 \\
Human metapneumovirus                             & 852 \\
Enterovirus A                                     & 834 \\
Human parainfluenza virus 1                       & 803 \\
HHV-8           & 792 \\
Human papillomavirus                              & 772 \\
Parechovirus A3                                   & 741 \\
Chikungunya virus                                 & 716 \\
Human coronavirus strain NL63                     & 696 \\
Vaccinia virus                                    & 696 \\
Human herpesvirus 7                               & 696 \\
Dengue virus                                      & 691 \\
Zika virus                                        & 669 \\
John Cunningham virus                             & 664 \\
Adenovirus                                        & 657 \\
Ebola virus                                       & 656 \\
Cytomegalovirus                                   & 654 \\
Hantavirus                                        & 648 \\
Rabies virus                                      & 648 \\
Hepatitis B virus                                 & 644 \\
Human herpesvirus 6                               & 644 \\
Varicella zoster virus                            & 639 \\
Herpes simplex virus 2                            & 635 \\
MERS coronavirus      & 633 \\
SFTS virus & 632 \\
Hepatitis A virus                                 & 621 \\
Enterovirus D                                     & 602 \\
Measles virus                                     & 600 \\
West Nile virus                                   & 600 \\
Parvovirus B19                                    & 600 \\
Human parainfluenza virus 2                       & 596 \\
Crimean-Congo hemorrhagic fever virus             & 590 \\
Herpes simplex virus 1                            & 588 \\
Hepatitis E virus                                 & 586 \\
Rubella virus                                     & 584 \\
Norovirus                                         & 579 \\
Human rhinovirus A                                & 578 \\
Molluscum contagiosum virus                       & 568 \\
Lassa virus                                       & 564 \\
Saffold virus                                     & 560 \\
Epstein-Barr virus                                & 545 \\
Enterovirus C                                     & 542 \\
Human rhinovirus B                                & 536 \\
Human Astrovirus                                  & 528 \\
Human coronavirus OC43                            & 520 \\
Respiratory syncytial virus                       & 474 \\
\bottomrule
\end{tabular}
\end{table}
\vspace*{\fill}
\clearpage

\section{LSTM Architecture}
\vspace*{\fill}
\begin{figure}[h]
\centering
\includegraphics[width=0.7\linewidth]{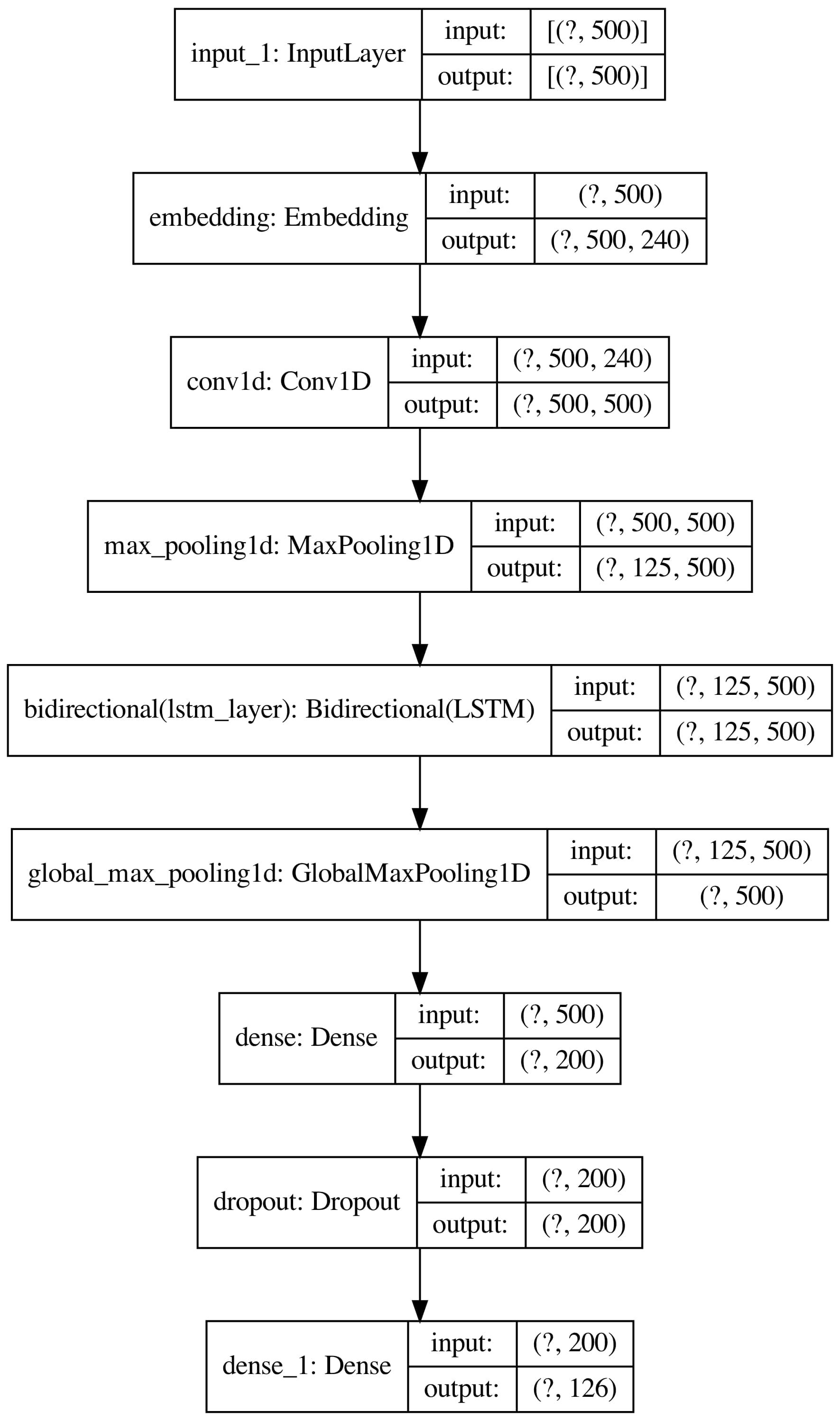}
\caption{LSTM architecture used. Number of trainable parameters: 1,740,266. "?" denotes the batch dimension. For more information on architecture selection, see \Cref{hyperparams}, Hyperparameter and Architecture Selection.}
\end{figure}
\vspace*{\fill}
\clearpage

\section{CNN Architecture}

\hvFloat[%
nonFloat=true,
objectAngle=90,%
capPos=after,%
capAngle=90,
capWidth=w,
floatCapSep=20pt]{figure}{
\includegraphics[scale=0.098]{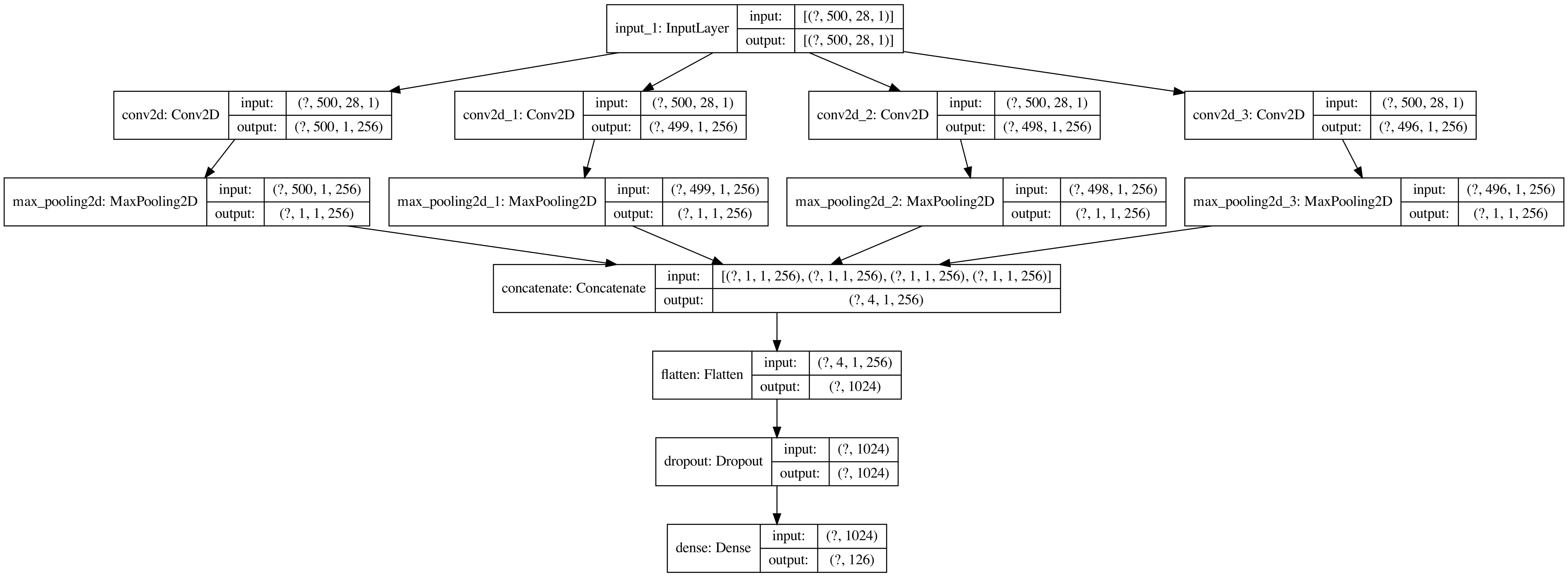}
}{CNN architecture used. Number of trainable parameters: 209,022. "?" denotes the batch dimension. For more information on architecture selection, see \Cref{hyperparams}, Hyperparameter and Architecture Selection.}{cnnimg}
\clearpage

\section{Hyperparameter and Architecture Selection}
\label{hyperparams}

\begin{table}[htbp]
\centering
\label{hyp_table}
\caption{List of hyperparameters experimented with. The hyperparameters used for generating the results in the article are highlighted in \textbf{bold}.}
\begin{tabular}{lcc}
\toprule
Hyperparameter         & LSTM                     & CNN                      \\
\midrule
Learning rate          & 1e-1, 1e-2, \textbf{1e-3}, 1e-4   & 1e-1, \textbf{1e-2}, 1e-3, 1e-4   \\
Batch size             & 32,  64, \textbf{128},  256,  512 & 32,  64, \textbf{128},  256,  512 \\
Threshold              & 0.9,  0.7,  0.5,  \textbf{0.2}, 0.1    & 0.9,  0.7,  0.5,  \textbf{0.2}    \\
Epochs                 & 5 to 30 \textbf{(20)}             & 5 to 30 \textbf{(20)}             \\
Sequence length cutoff & \textbf{500}, 1000                & \textbf{500}, 1000      \\        
\bottomrule
\end{tabular}
\end{table}
The hyperparameters tested in our experiments are presented in \Cref{hyp_table} It is certainly possible to improve the accuracies of our experiments by conducting a vaster coverage of the loss landscape through more extensive training (e.g. running longer experiments with smaller learning rates on more complex network architectures), especially for results in \textit{Experiment II}. However, due to performance constraints, the scope of hyperparameter tuning as well as the ANN architectures experimented on are relatively constrained as we focused on the methodology as opposed to optimal performance in this study. It should be noted that much improvement is possible in this front, as pointed out in Discussion and Future Work. Additional notes regarding our observations during hyperparameter tuning are presented below.
\begin{itemize}
    \item For the threshold, we wanted to predict eagerly, i.e. we considered false negatives more costly errors than false positives. A high threshold would mean the outputs would be composed only of the antivirals our models are very confident about per amino acid sequence. This we deem undesirable, as while we do hope these outputs narrow the scope of antivirals to focus on, over-restricting could prevent antivirals that are predicted frequently yet with low probability be detected. A low threshold such as 0.2 filtered the number of antivirals sufficiently, but also left enough breathing room for the domain experts to draw their own conclusions on a per-drug basis.
    \item While a larger sequence length cutoff was possible and not detrimental to the results, we deemed 500 a suitable trade-off in terms of performance versus accuracy, as many sequences do not reach lengths in the thousands to begin with.
    \item As mentioned, the number of epochs trained could be increased, as we did not see dramatic signs of overfitting at 20 epochs or further. However, a flattening of the metrics were evident around 20 epochs with the hyperparameters listed, which therefore was selected a suitable stopping point.
\end{itemize}

\clearpage

\section{Output table sample}
\vspace*{\fill}
\hvFloat[%
nonFloat=true,
objectAngle=90,%
capPos=r,%
capAngle=90,
capWidth=h,
floatCapSep=10]{table}{
\begin{tabular}{@{}lllll@{}}
\toprule
FASTA Sequence & Virus Name                  & GenBank\_Title          & Antivirals & Probabilities \\ \midrule
MKFLVFLGII     & SARS-CoV-2 & ORF8 protein           & Rapamycin, Mitoxantrone              & 0.994, 0.351                                           \\
MGYINVFAFPFT   & SARS-CoV-2 & ORF10 protein, partial & Brincidofovir, Nitazoxanide, Tilorone & 0.959, 0.865, 0.452           \\
MESLVPGFNE     & SARS-CoV-2 & orf1ab polyprotein     & Lopinavir, Ritonavir, Tilorone,       & 0.761, 0.653, 0.280           \\ \bottomrule
\end{tabular}
}{A section of sample outputs for amino acid sequences and their associated antivirals. Post-processing outputs a list of drugs that were selected along with the respective probabilities of the drugs being “effective” against the virus with the given amino acid sequence.}{outdb}
\vspace*{\fill}
\clearpage

\section{Filter Activation Visualizations}

\begin{figure}[!ht]
\centering
\includegraphics[width=\linewidth]{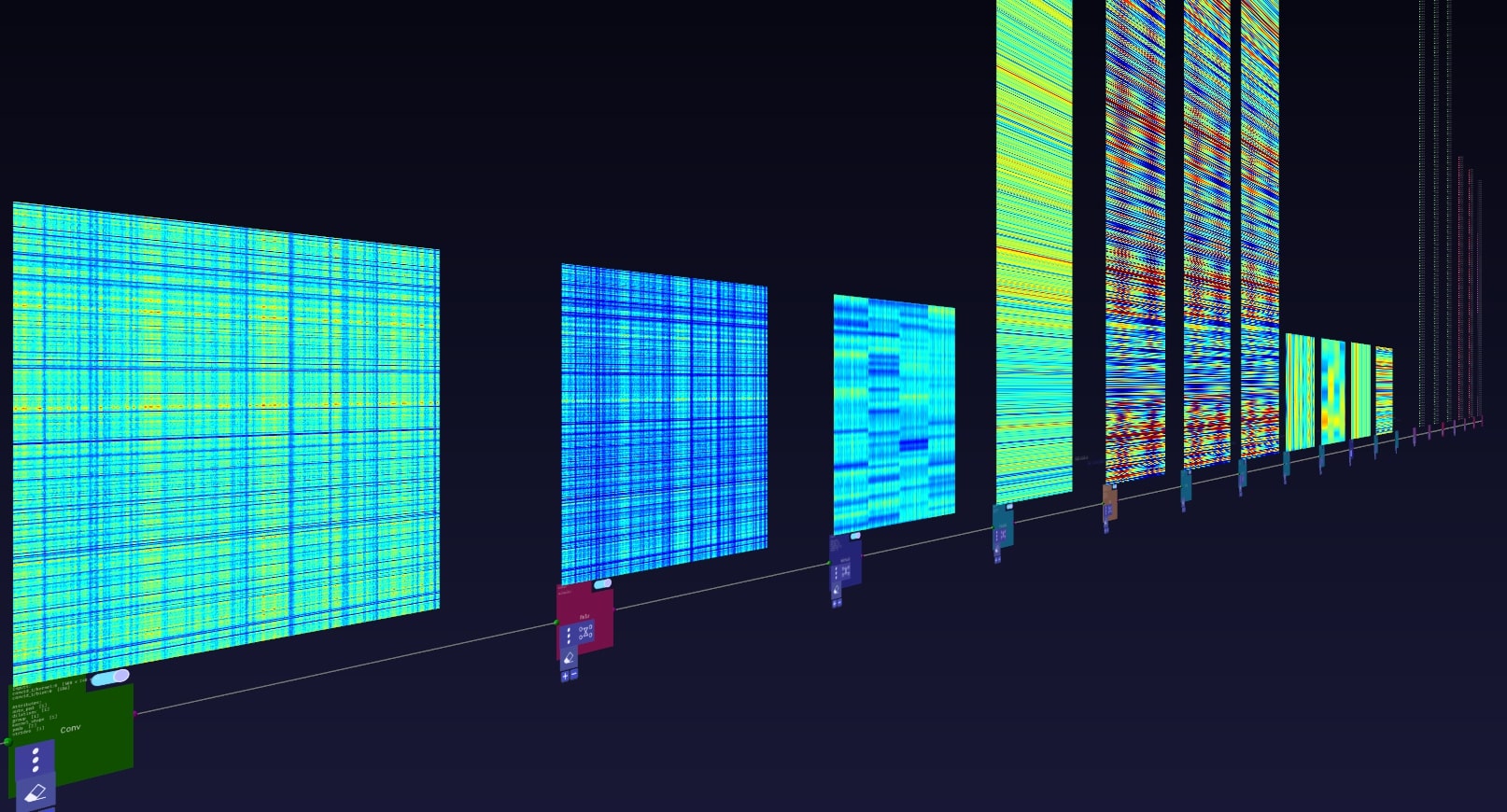}
\caption{Activation visualizations of the LSTM layers for a given sequence}
\end{figure}

\begin{figure}
     \centering
     \begin{subfigure}[b]{0.29\textwidth}
         \centering
         \includegraphics[width=\textwidth]{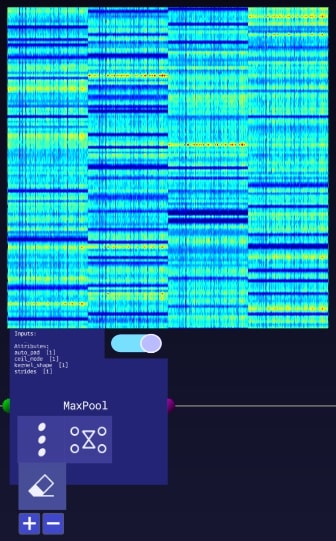}
         \caption{MaxPool}
     \end{subfigure}
     \hfill
     \begin{subfigure}[b]{0.3\textwidth}
         \centering
         \includegraphics[width=\textwidth]{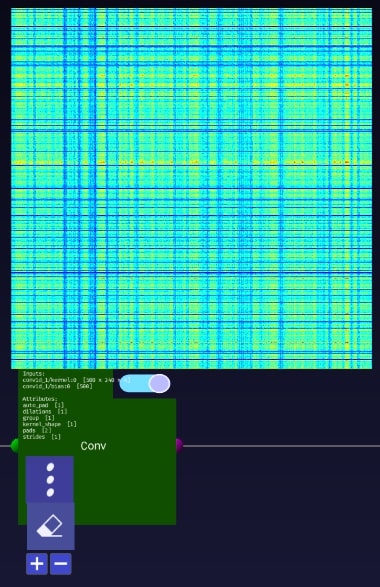}
         \caption{Conv1D}
     \end{subfigure}
     \hfill
     \begin{subfigure}[b]{0.29\textwidth}
         \centering
         \includegraphics[width=\textwidth]{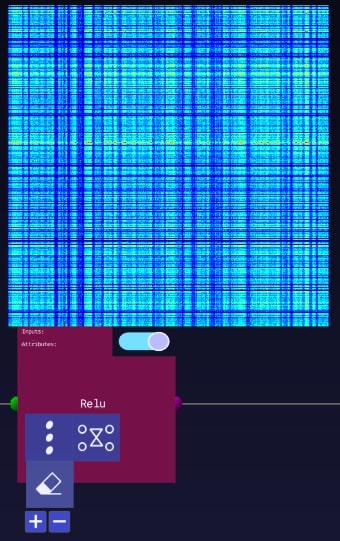}
         \caption{ReLU}
     \end{subfigure}
        \caption{Filter activations for three different LSTM layers for a given amino acid sequence. Warmer colors indicate higher activation regions.}
\end{figure}

\begin{figure}
     \centering
     \begin{subfigure}[b]{0.223\textwidth}
         \centering
         \includegraphics[width=\textwidth]{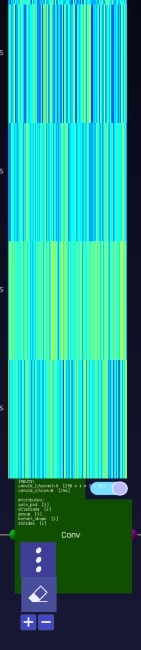}
         \caption{Conv2D}
     \end{subfigure}
     \hfill
     \begin{subfigure}[b]{0.211\textwidth}
         \centering
         \includegraphics[width=\textwidth]{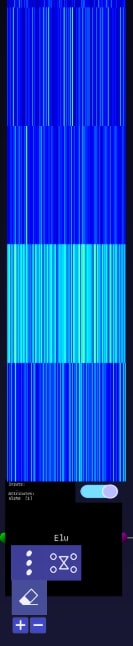}
         \caption{Elu}
     \end{subfigure}
     \hfill
     \begin{subfigure}[b]{0.228\textwidth}
         \centering
         \includegraphics[width=\textwidth]{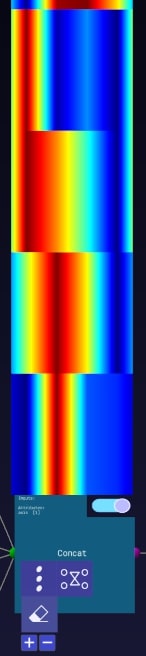}
         \caption{Concat}
     \end{subfigure}
     \hfill
     \begin{subfigure}[b]{0.221\textwidth}
         \centering
         \includegraphics[width=\textwidth]{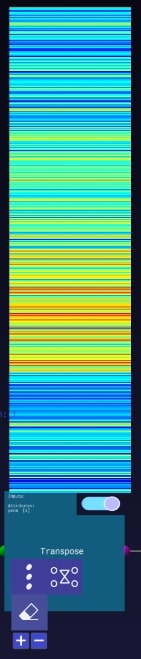}
         \caption{Transpose}
     \end{subfigure}
        \caption{Filter activations for four different CNN layers for a given amino acid sequence. Warmer colors indicate higher activation regions.}
\end{figure}

\end{appendices}

\end{document}